\pdfoutput=1
\documentclass[11pt]{article}
\usepackage[preprint]{acl}
\newcommand{\dochplt}{DocHPLT\xspace}
\newcommand{\hplt}{HPLT\xspace}
\usepackage{times}
\usepackage{amsmath}
\usepackage[T1]{fontenc}
\usepackage[utf8]{inputenc}
\usepackage{microtype}
\usepackage{amssymb}
\usepackage{pifont}
\usepackage{comment}
\usepackage{inconsolata}
\usepackage{graphicx}
\usepackage{tcolorbox}
\usepackage{booktabs}
\usepackage{enumitem}
\usepackage{multicol}
\usepackage{multirow}
\usepackage{makecell}
\usepackage{xspace}
\usepackage{makecell}
\usepackage{placeins}
\usepackage{lettrine}
\title{\dochplt: A Massively Multilingual Document-Level Translation Dataset}
\author{
  Dayyán O'Brien\textsuperscript{\raisebox{0.1ex}{$\bigstar$},\raisebox{-0.3ex}{\includegraphics[width=2.25ex]{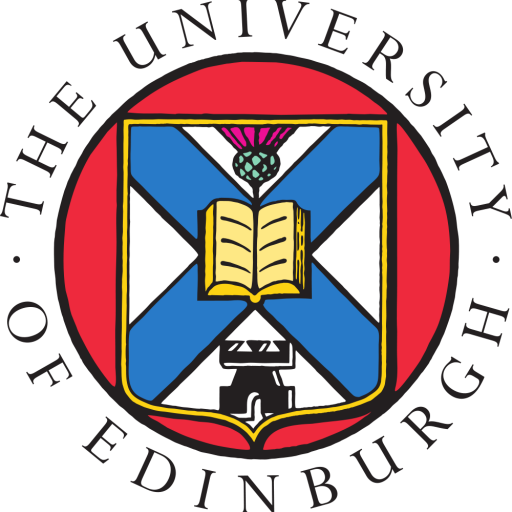}}} \qquad
  Bhavitvya Malik\textsuperscript{\raisebox{0.1ex}{$\bigstar$},\raisebox{-0.3ex}{\includegraphics[width=2.25ex]{figures/edinburgh.png}}} \qquad
  Ona de Gibert\textsuperscript{\raisebox{-0.1ex}{\includegraphics[width=2.5ex]{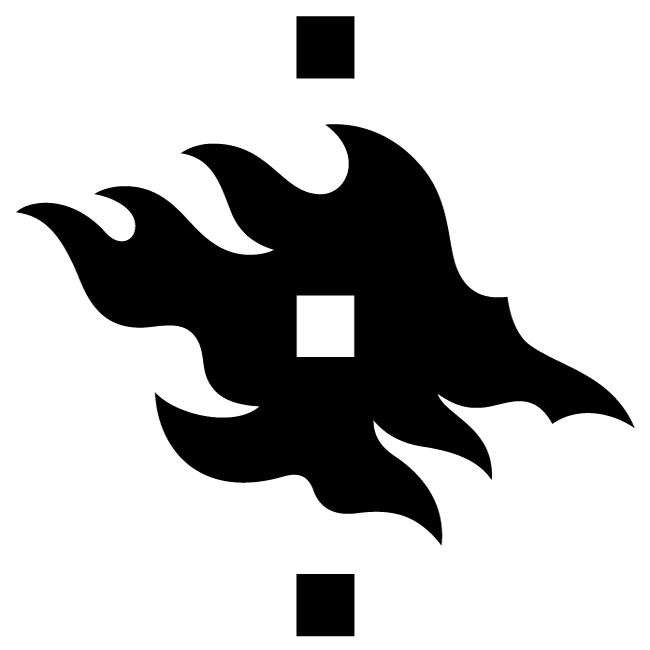}}} \\
  \textbf{Pinzhen Chen}\textsuperscript{\raisebox{-0.3ex}{\includegraphics[width=2.25ex]{figures/edinburgh.png}}} \qquad
  \textbf{Barry Haddow}\textsuperscript{\raisebox{-0.3ex}{\includegraphics[width=2.25ex]{figures/edinburgh.png}}} \qquad
  \textbf{Jörg Tiedemann}\textsuperscript{\raisebox{-0.1ex}{\includegraphics[width=2.5ex]{figures/helsinki.png}}} \\
  \textsuperscript{\raisebox{-0.3ex}{\includegraphics[width=3ex]{figures/edinburgh.png}}}University of Edinburgh \qquad
  \textsuperscript{\raisebox{-0.1ex}{\includegraphics[width=3ex]{figures/helsinki.png}}}University of Helsinki  \\
  \texttt{\{dayyan.obrien,bmalik2,pinzhen.chen,bhaddow\}@ed.ac.uk} \\
  \texttt{\{ona.degibert,jorg.tiedemann\}@helsinki.fi}
}
\begin{document}
\maketitle
\def\thefootnote{$\bigstar$}\footnotetext{Equal contribution. Public access to \dochplt{}: \url{https://huggingface.co/datasets/HPLT/DocHPLT}.}
\def\thefootnote{\arabic{footnote}}
\begin{abstract}
Existing document-level machine translation resources are only available for a handful of languages, mostly high-resourced ones. To facilitate the training and evaluation of document-level translation and, more broadly, long-context modeling for global communities, we create \dochplt, the largest publicly available document-level translation dataset to date. It contains 124 million aligned document pairs across 50 languages paired with English, comprising 4.26 billion sentences. By adding pivoted alignments, practitioners can obtain 2500 additional pairs not involving English. Unlike previous reconstruction-based approaches that piece together documents from sentence-level data, we modify an existing web extraction pipeline to preserve complete document integrity from the source, retaining all content, including unaligned portions. After our preliminary experiments identify the optimal training context strategy for document-level translation, we demonstrate that LLMs fine-tuned on \dochplt{} substantially outperform off-the-shelf instruction-tuned baselines, with particularly dramatic improvements for under-resourced languages. We open-source the dataset under a permissive license, providing essential infrastructure for advancing multilingual document-level translation.
\end{abstract}
\section{Introduction}
The field of natural language processing (NLP) is shifting its focus toward end-to-end, complex tasks, including the domain of machine translation. This increases the demand for techniques and resources beyond the sentence level, with document-level machine translation (DocMT) being a prime example \citep{maruf-haffari-2018-document,zhang-etal-2018-improving,agrawal-etal-2018-contextual,huo-etal-2020-diving}. While there is not a single definition of a document, DocMT requires models to translate more than one sentence as a coherent unit rather than isolated segments. This approach is necessary for handling various discourse phenomena: \textit{anaphora, deixis, ellipsis, discourse connectives, grammatical and lexical cohesion} \citep{10.1145/3441691}, which sentence-level translation typically loses \citep{muller-etal-2018-large,bawden-etal-2018-evaluating,voita-etal-2018-context}. Recent long-context large language models (LLMs) are well-suited for this task, as they are usually pre-trained to process thousands of tokens at a time. However, DocMT remains largely unexplored or untested for most languages due to a simple but significant problem: we lack document-level parallel data for both model building and evaluation.

Historically, parallel corpora were mostly constructed in a sentence-oriented manner, using a pipeline that split the text into sentences, aligned them, and then discarded unaligned and multiply-aligned sentences.   
While a handful of language pairs have some document-level MT resources, the majority of languages have none. This creates two related problems at once: we cannot build DocMT for these languages, and we cannot evaluate DocMT properly. As NLP research moves toward more end-to-end, context-aware applications, this data gap means that most languages get left behind.

We tackle this problem by extracting parallel documents from large web crawls, but our methodology differs from the majority of previous efforts that reconstruct data from sentence pairs after the fact. Instead, we modify the web extraction pipeline itself to preserve document structure from the beginning, retaining documents in their entirety with all original context and non-parallel text. For each language pair, we deliver the aligned documents along with quality-scored sentence alignments and alignment density metrics.

Our effort yielded \dochplt, a large multilingual document-level translation dataset covering 50 language pairs with English, listed in Appendix~\ref{appen:data_stats}. The resulting corpus contains 87.8 million documents in English and 50 other languages, 124 million aligned document pairs, and 4.26 billion sentences. A highlight of our work is the focus on medium- and low-resource languages that previous DocMT datasets have overlooked. Practitioners can also use English as a pivot to align up to 2500 extra non-English pairs, expanding the dataset's usefulness beyond English-centric translation.

Using \dochplt, we conduct extensive experiments with different modeling methods for LLM-based document-level translation. We first try different context sizes for LLM fine-tuning: 1) full document-to-document training with loss calculated on the entire target; and 2) chunk-based training with loss computed on individual segments. These experiments determine the optimal context granularity for our subsequent work. Then, in addition to prompting off-the-shelf instruction-tuned large language models (LLMs) as a baseline, we run monolingual and multilingual fine-tuning using \dochplt{}, tested on both seen and unseen languages. The usefulness of our data is reflected empirically: LLMs fine-tuned on our data consistently outperform prompting baselines, showing that practitioners can gain strong performance in DocMT for languages often considered ``unsupported'' in the machine translation research community. 

In summary, our contributions, centred around the \dochplt{} resource, are as follows:
\begin{itemize}
    \item \textbf{Scale and diversity:} \dochplt is the \textbf{largest} publicly available document-level translation resource: 124M document pairs for 50 languages paired with English, totaling 4.26B sentences, with extensive medium- and low-resource coverage.
    \item \textbf{Document-first approach}: Instead of piecing together documents from aligned sentence pairs, we preserve complete documents with original structure and unaligned text, enriched with quality metrics such as alignment density and sentence pair-level scores.
    \item \textbf{Empirical validation}: Through LLM experiments on both the internal test set and WMT24++, we establish baselines, test different training strategies, and demonstrate gains in DocMT using our data.
\end{itemize}

\section{Related Work}
\subsection{Document-Level Translation} 
Document-level translation aims to process an entire document as a coherent unit, rather than processing each sentence independently. 
This paradigm leverages the ability of modern neural architectures, lately LLMs, to handle long context, making it particularly effective for capturing document-level discourse structures. Recent work has demonstrated that going beyond sentence-level translation is essential for handling discourse phenomena such as coreference resolution \citep{muller-etal-2018-large,bawden-etal-2018-evaluating,voita-etal-2018-context}. The development of dedicated document-level benchmarks further reflects this growing interest in evaluating MT systems in context \citep{guillou-hardmeier-2016-protest,jwalapuram2020can,wicks-post-2023-identifying,fernandes-etal-2023-translation}.

Moreover, a variety of modeling strategies have been proposed for DocMT \citep{tiedemann-scherrer-2017-neural,maruf-haffari-2018-document,zhang-etal-2018-improving,agrawal-etal-2018-contextual,sun-etal-2022-rethinking}, and more recent works adapt LLM-based architectures \citep{wang-etal-2023-document-level,petrick-etal-2023-document,wu2024adapting,jin2024chaptertochaptercontextawareliterarytranslation,ramos2025multilingual,hu2025source}. However, there is still no standard practice in training to ensure effective context handling or in assessing document-level translation. Also, performance gains over strong sentence-level baselines remain inconsistent and not clearly attributable to effective context utilization \citep{kim-etal-2019-document}. In this work, we try out various context sizes in LLM fine-tuning to establish effective training strategies on \dochplt{}.

\subsection{Document-Level Translation Data}
Although there have been several massive-scale parallel corpus mining efforts \citep{banon-etal-2020-paracrawl,el-kishky-etal-2020-ccaligned,schwenk-etal-2021-ccmatrix,de-gibert-etal-2024-new,burchell-etal-2025-expanded}, document-level data remain limited in size and scope, particularly when extending beyond English-centric or high-resource languages. Moreover, there is little agreement on what constitutes a ``document''; definitions vary widely across studies, ranging from short paragraphs to entire articles or books. This lack of standardization, combined with the scarcity of large-scale multilingual document-level corpora, motivates the need for more diverse resources such as the one we present in this work. Before illustrating our data methodology, we survey two typical methods used in creating document-level translation data.

\paragraph{Reconstruction-based} A common strategy to obtain document-level parallel data is to reconstruct from existing sentence-level data. Notably, the sentence-level ParaCrawl \citep{banon-etal-2020-paracrawl} has been widely used as a seed for this purpose. \citet{al-ghussin-etal-2023-exploring} extracted English–German parallel paragraphs from ParaCrawl, although these are not full-document units. ParaDocs \citep{wicks-etal-2024-recovering} recovered document-level data from ParaCrawl, News Commentary \citep{kocmi-etal-2023-findings}, and Europarl \citep{koehn-2005-europarl} for German, French, Spanish, Italian, Polish, and Portuguese, all paired with English. Similarly, \citet{pal-etal-2024-document} released a large-scale reconstructed corpus for German, French, Czech, Polish, and Russian---again all paired with English, along with an open-source pipeline for extension to other languages. 

\paragraph{Collection-based} An alternative approach is to collect or create document-level parallel corpora directly from targeted sources from scratch. Earlier efforts include Europarl based on the proceedings of the European Parliament \citep{koehn-2005-europarl} and OpenSubtitles from movie and TV subtitles \citep{lison-tiedemann-2016-opensubtitles2016}, but these essentially consist of ``spoken'' documents, where the former is divided into speeches and the latter into films/shows. Literary works have also been a popular origin. \citet{jiang2022bilingual} introduced a Chinese–English corpus based on web novels, where each chapter, with a median of 30 sentences, is treated as a document. 
\citet{thai-etal-2022-exploring} created PAR3 by aligning machine and human translations of 118 novels across 19 languages at the paragraph level. \citet{jin2024chaptertochaptercontextawareliterarytranslation} constructed JAM from 160 English-Chinese novel pairs with chapter-level alignment. 
More recently, \citet{alabi2025afridoc} created AFRIDOC-MT, a document-level translation corpus sourced from IT news and health articles and manually translated from English. By covering Amharic, Hausa, Swahili, Yorùbá, and Zulu, it extends DocMT data to medium and lower-resourced languages. \citet{wastl-et-al-2025-20min} scraped a Swiss online news outlet to create 20min-XD, a French-German dataset. Such data, directly derived from resources intended to be document-aligned, is high-quality but often limited by languages due to the coverage of the upstream source and/or the cost and effort required.

\paragraph{Key methodological differences in this work} 
Our work gathers document translations from large web crawls, but differs fundamentally from reconstruction-based approaches. As explained later in Section~\ref{sec:data-methodology}, rather than piecing together documents from sentence pairs post-hoc, we modify the document alignment stage of the extraction pipeline to preserve complete document structure from the beginning. This document-first methodology ensures we retain all original content, including unaligned portions, positioning our approach as collection-based at the document level while leveraging existing text crawling and processing infrastructure.

\section{\dochplt}\label{sec:data-methodology}
In this section, we explain how we modify and then apply an existing parallel sentence extraction pipeline from ParaCrawl to extract a document-level corpus from a large multilingual web crawl, HPLT.

\begin{figure*}[t]
    \centering\small
    \includegraphics[clip, trim=0cm 18.2cm 0cm 0.8cm, width=0.75\linewidth]{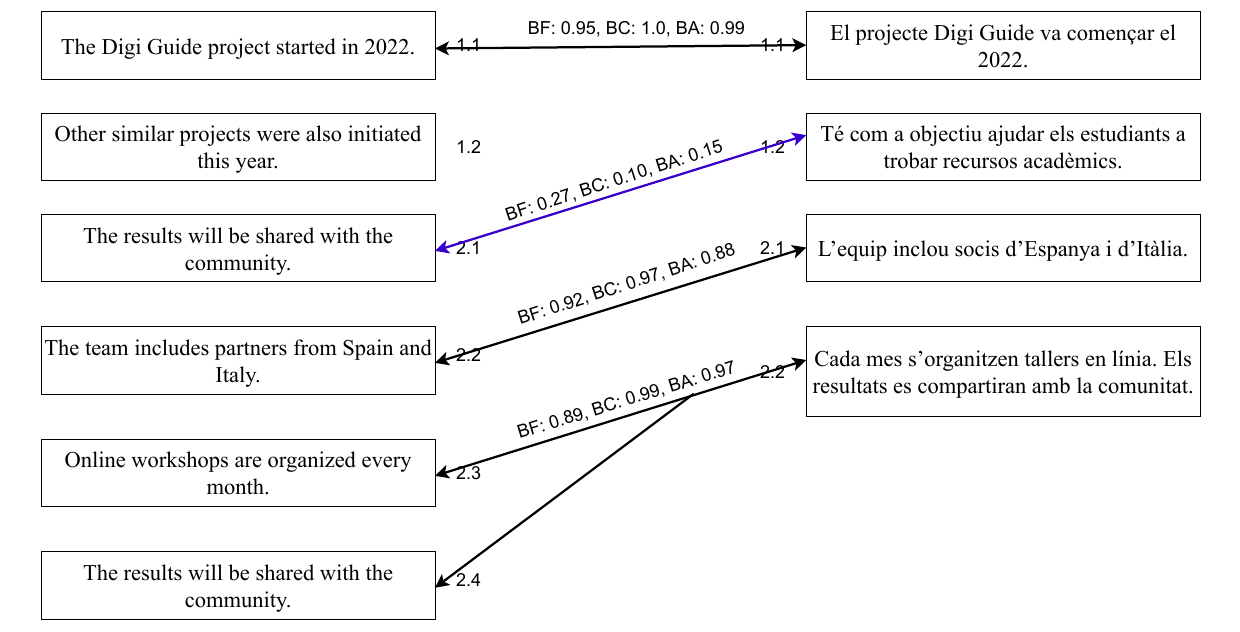}
    \\\footnotesize \textbf{BF}: Bifixer, \textbf{BC}: Bicleaner, \textbf{BA}: BLEUalign. Higher values indicate better alignment and translation quality.
    \caption{An example of good, bad (in blue), and multi-way alignments for English-Catalan docs.}
    \label{fig:alignment_example}
\end{figure*}

\subsection{Dataset Creation}

The starting point for our dataset creation is 15TB of cleaned web documents derived from the Internet Archive\footnote{\url{https://archive.org/}} and CommonCrawl\footnote{\url{https://commoncrawl.org/}} released as version 2 of the  \hplt corpus \citep{burchell-etal-2025-expanded}.
In the preparation of \hplt, the document text was extracted from HTML using Trafilatura \citep{barbaresi-2021-trafilatura} and language-classified using openLID \citep{burchell-etal-2023-open}. In this work, a \textit{document} is defined as \textit{the full text content retrieved from archive snapshots of a specific URL}.

To extract a parallel corpus of documents, we use a modified version of the  ParaCrawl extraction pipeline \citep{banon-etal-2020-paracrawl}.
The original pipeline is sentence-oriented, i.e.\ it produces a sentence-aligned corpus and discards unaligned sentences. But because the pipeline runs document alignment followed by sentence alignment in separate stages, we are able to intervene to produce document-oriented data. We extract and record each pair of aligned documents, then map the unfiltered sentence alignments back into their source documents. This document-first methodology ensures we retain all document content, even unaligned portions, to provide richer context than traditional parallel corpora.

\paragraph{Document structuring} 
We transform each document into a hierarchical XML representation that preserves its internal structure. Paragraphs are split by newline characters and maintain their original boundaries, while the text of each paragraph is segmented into sentences using the Loomchild Segmenter \citep{loomchild}. Every structural element receives a unique identifier with paragraphs, such as \texttt{<P id="4">}, and sentences, such as \texttt{<s id="4.3">}. This structured representation allows us to track alignments at both document and sentence levels while retaining all content from the original \hplt documents.

\paragraph{Content-based deduplication}
Since our initial collection contains multiple temporal snapshots of the same URLs, we implement a content-based deduplication strategy. First, within each language-specific collection, we remove duplicates, using the URL together with the full text as the key. This ensures we keep only unique document versions for each URL. Second, we perform global deduplication, based on the URL and text as the key, across all English documents from the 50 language pairs, consolidating them into a single collection. This is necessary because the same English document may appear in multiple language pairs (e.g., the same English page aligned to both Basque and Catalan translations). After deduplication, we have a clean collection of unique documents for each of the 50 source languages and a single, unified collection for all English documents, ensuring each unique document version appears exactly once while preserving all alignment relationships.

We deliberately preserve duplicate content across different URLs and retain near-duplicates within the same URL. This design choice maximizes research flexibility by allowing downstream users to apply filtering strategies suited to their particular use cases. Additionally, duplicate content from different URLs preserves valuable metadata, particularly the source URL, which may indicate different domains, publication contexts, or content distribution patterns. Near-duplicates also represent meaningful content variations such as updates, revisions, or editorial differences. Deduplicating content only for each language separately results in a 3.3\% drop in our document count from the original \dochplt{} corpus (see Appendix \ref{appen:data_stats}).

\paragraph{Alignment verification and generation}
The ParaCrawl pipeline originally used MinHash \citep{minhash} to deduplicate similar sentences, grouping them and assigning the same quality scores regardless of their source documents. We modify this step to remove MinHash deduplication entirely, instead maintaining all original texts and tracking which documents they came from. This allows us to preserve the complete document structure while still computing alignment quality scores---BLEUalign \citep{sennrich-volk-2010-mt}, Bicleaner, and Bifixer \citep{ramirez-sanchez-etal-2020-bifixer}---for each sentence pair. The document in Figure~\ref{fig:alignment_example} illustrates examples of good, bad, and multi-way alignments along with their corresponding quality scores. We then map each alignment back to its specific source and target documents, maintaining the document-sentence relationships throughout the process. For any document that had multiple versions with the same URL, we explicitly check that every sentence referenced in an alignment link actually exists in the final XML file to ensure it references the correct version(s). The output follows the standard \texttt{cesAlign} XML format\footnote{\url{https://opus.nlpl.eu/legacy/trac/wiki/DataFormats.html}}, where each alignment links specific sentence IDs between source and target documents along with their quality scores. 

\paragraph{MultiDocHPLT by pivoting via English}
As a ``bonus'' data release, English can be used as a pivot language to derive a corpus beyond the English-centric pairs. This enables the modeling and evaluation of DocMT between two non-English languages. The process is straightforward: if a document in a language and another document in another language are both aligned to the same English document, then we assume a direct alignment between the two documents. We pivot the sentence alignments in a similar way.

\subsection{Data Statistics} 
In this section, we present the statistics of our English-centric dataset. We provide full tables in Appendix~\ref{appen:data_stats}: Table~\ref{tab:language_totals} details total documents and sentences per language, while Table~\ref{tab:alignment_stats} reports alignment statistics for each language pair. Specifically, for each language pair, we report the number of aligned document pairs (\#doc pairs), the total number of alignments  (\#alignments), the average number of sentences per aligned document (avg \#aligns./\#docs), document length ratio calculated as the average number of sentences in English relative to the target language (avg \#sent\_en/\#sent\_xx), number of sentences per document (\#sent/\#docs), and the average alignment scores.

Across all language pairs, \dochplt{} contains 87.8 million unique documents with 4.26 billion total sentences, averaging 48.6 sentences per document. The English collection dominates with 47.5 million documents (2.67 billion sentences), while individual non-English languages range from Japanese with 4 million documents (164 million sentences) down to Xhosa with 22 thousand documents (996 thousand sentences). These documents form 124 million aligned document pairs, with an average of 14.8 sentence-level alignments per document pair. Document coverage varies significantly: Japanese-English and Turkish-English each contribute over 11 million aligned document pairs, respectively, while under-represented languages like Sinhala-English (123 thousand document pairs), Uzbek-English (157 thousand document pairs), and Xhosa-English (44 thousand document pairs) have substantially smaller collections.

We observe considerable variations in document length ratios between aligned pairs, ranging from 3.91 (Malayalam-English), where the English documents are typically longer, to 0.84 (Arabic-English). 
Additionally, the average Bicleaner scores vary significantly, with language pairs like Arabic-English (0.700) demonstrating relatively high-quality alignments, whereas pairs such as Maltese-English (0.293) display substantially lower average alignment quality.

\paragraph{Alignment density}
Furthermore, we calculate alignment density (AD), which is defined as the proportion of aligned sentence pairs between two documents relative to the length of the longer document.  Formally, given two documents \(D_{\text{src}}\) and \(D_{\text{tgt}}\), with \(|D_{\text{src}}|\) and \(|D_{\text{tgt}}|\) denoting their respective sentence counts, the alignment density  is computed as
\begin{equation}
AD = \frac{\text{\# of aligned sentence pairs}}{\max(|D_{\text{src}}|, |D_{\text{tgt}}|)}
\end{equation}
Alignment density ranges between 0 (no aligned pairs) and 1 (perfect sentence-level coverage) if alignments are strictly one-to-one; however, since our alignment procedure allows one-to-many and many-to-one mappings, values above 1 are also possible. 
This feature may reveal the quality and the characteristics of the documents: an AD of exactly 1 could suggest that the documents were machine-translated (at the sentence level), whereas a very low AD might imply that they were accidentally matched, possibly due to high-frequency phrases or placeholders. 

We observe considerable variation in AD across language pairs, e.g., Welsh-English (cy-en) and Afrikaans-English (af-en) show notably high average alignment densities (0.426 and 0.446, respectively), compared to languages like Farsi, Malayalam, and Marathi, where alignments are much sparser (0.153, 0.151, and 0.150, respectively). While some language pairs exhibit higher or lower densities, these scores are better understood relative to other scores rather than absolute terms.

We did not observe any consistent correlation between automatic quality metrics (BicleanerAI and CometKiwi) and AD values. Future work should investigate more carefully how AD should be interpreted and framed.

\section{Experiments and Findings}
In order to test the usefulness of our dataset, we apply it to the task of fine-tuning  LLMs for MT. Our first set of experiments tests different context lengths for this fine-tuning to see how much performance is affected by using the larger document contexts that \dochplt{} enables.  We then compare this best-performing fine-tuned configuration against off-the-shelf instruction-tuned models on the same test set. Finally, we investigate monolingual and multilingual fine-tuning for DocMT on \dochplt{}. In the following sections, the notation of ``src-trg'' refers to the src-to-trg translation direction.

\paragraph{Languages}
We test translation from English into a total of 10 languages, chosen for their diversity in script, typology, and resource availability, as well as their inclusion in WMT24++ \citep{deutsch-etal-2025-wmt24} (for testing) and CometKiwi \citep{rei-etal-2022-cometkiwi} (for filtering). The languages are Arabic (ar), Catalan (ca), Hindi (hi), Estonian (et), Persian (fa), Finnish (fi), Icelandic (is), Korean (kr), Malayalam (ml), and Urdu (ur). It is worth noting that generating non-English is usually harder for LLMs compared to generating English.

\paragraph{Data processing} We preprocess the documents for training by removing those with an AD below 0.3 or a document-averaged Bicleaner score below 0.3. We then discard unaligned segments in either source or target, and merge segments in a one-to-many alignment into a single segment. Finally, we filter data using CometKiwi \citep{rei-etal-2022-cometkiwi} with SLIDE \citep{raunak-etal-2023-evaluating}: a window of 3 and a slide of 1. We retain document pairs with a CometKiwi score in the top 25\textsuperscript{th} percentile for every language. This is to ensure that only high-quality parallel documents are used for training or evaluation.

\paragraph{Model training and inference} We perform supervised fine-tuning (SFT) on Qwen2.5-7B-Instruct \citep{qwen2.5} and Llama-3.1-8B-Instruct \citep{grattafiori2024llama3herdmodels} with LoRA, rank 16 and alpha 32 \citep{hu2021loralowrankadaptationlarge}, using the \texttt{open-instruct} toolkit\footnote{https://github.com/allenai/open-instruct}. Unless stated otherwise, our models are fine-tuned on 1000 documents per language, due to compute constraints. Our hyperparameters are listed in Appendix~\ref{appen:hyperparam}. 
At test time, we always translate an entire source document in a single pass. LLM's chat template is always applied. All prompt information is detailed in Appendix~\ref{appen:prompts}.

\paragraph{Evaluation set} We conduct evaluations on two test sets: a held-out set from \dochplt{} and WMT24++, selected for their overlapping language coverage. We construct \dochplt{} test by randomly sampling 500 documents per language from the CometKiwi-filtered corpora. We de-contaminate on the English side by computing Jaccard similarity over bigrams and removing any test document with a similarity above 0.8 to any training document. This ensures that our evaluation is not biased towards training on \textit{similar} documents. The final test sizes are shown in Table \ref{tab:test_set}.

\begin{table}[t]
\centering\small
\begin{tabular}{lc}
\toprule
& \textbf{\#test docs} \\
\midrule
en-fi & 489  \\
en-is & 492  \\
en-ko & 497 \\
en-ml & 473  \\
en-ur & 486 \\
\bottomrule
\end{tabular}
\caption{Test sizes after de-near-duplication (de-contamination); always 500 for unseen language pairs.}
\label{tab:test_set}
\end{table}

\paragraph{Metrics}  We compute BLEU\footnote{nrefs:1|case:mixed|eff:no|tok:13a|smooth:exp|version:2.5.1} \citep{papineni-etal-2002-bleu} and chrF++\footnote{nrefs:1|case:mixed|eff:yes|nc:6|nw:2|space:no|version:2.5.1} \citep{popovic-2017-chrf} by treating each hypothesis document and reference document as a single string, and then averaging these scores across all documents. Our metric choice avoids the need for sentence-level alignment, which DocMT outputs do not guarantee. We note that while neural metrics such as COMET or LLM-as-a-judge are generally more reliable at the sentence level, their effectiveness in our document-level setting remains uncertain due to limited empirical validation, context support, and language coverage.

\begin{table*}[t]
\centering\small
\begin{tabular}{rrrrrrrrrrrr}
\toprule
 &  \multirow[b]{2}{*}{\makecell{FT chunk\\(num sent)}} & \multicolumn{5}{c}{\dochplt{}}  & \multicolumn{5}{c}{WMT24++} \\ \cmidrule(lr){3-7}\cmidrule(lr){8-12}
 &  & en-fi  & en-is  & en-ko  & en-ml  & en-ur  & en-fi  & en-is  & en-ko  & en-ml  & en-ur  \\ 
\midrule
\multirow{10}{*}{\makecell{Qwen2.5-\\7B-Instruct}}  & 1 & 8.39  & 20.13  & 10.39  & \textbf{13.97} & 11.05  & 11.49  & \textbf{10.50} & 5.68  & \textbf{4.63}  & \textbf{9.16}  \\
 & 2 & 12.39  & 18.92  & 15.07  & 12.47  & \textbf{11.48} & \textbf{12.81} & 9.67  & 6.21  & 4.19  & 8.48  \\
 & BLEU\hspace{6ex}\hfill5 & 12.80  & 28.99  & 22.69  & 10.20  & 8.94  & 12.38  & 9.55  & \textbf{6.60}  & 2.93  & 5.72  \\
 & 10  & \textbf{13.87} & \textbf{32.54} & \textbf{23.71} & 12.20  & 11.11  & 12.20  & 9.27  & 6.37  & 3.67  & 5.75  \\
 & doc2doc  & 8.35  & 24.65  & 15.57  & 9.35  & 5.05  & 9.40  & 6.02  & 6.02  & 1.21  & 2.44  \\ \cmidrule(lr){2-12}
 & 1 & 30.51  & 40.49  & 23.84  & \textbf{38.72} & 32.76  & 36.37  & \textbf{32.71} & 19.36  & \textbf{29.46} & \textbf{31.98} \\
 & 2 & 39.61  & 39.02  & 30.80  & 37.73  & 34.66  & 39.05  & 31.62  & 20.40  & 27.89  & 30.55  \\
 & chrF++\hspace{6ex}\hfill5 & 42.12  & 50.00  & 39.24  & 33.05  & 30.71  & \textbf{39.53} & 31.30  & \textbf{21.27} & 22.90  & 25.49  \\
 & 10  & \textbf{44.01} & \textbf{53.74} & \textbf{40.87} & 37.23  & \textbf{34.72} & 38.71  & 30.46  & 20.90  & 26.65  & 26.37  \\
 & doc2doc  & 35.12  & 46.10  & 30.70  & 32.60  & 24.05  & 34.25  & 24.59  & 19.30  & 16.67  & 16.42  \\
\midrule
\multirow{10}{*}{\makecell{Llama-3.1-\\8B-Instruct}} & 1 & 7.23  & 6.51  & 6.36  & 16.16  & 12.54  & 8.53  & 6.06  & 2.24  & \textbf{4.32}  & 9.56  \\
 & 2 & 10.49  & 11.53  & 10.98  & \textbf{16.37} & 14.11  & 11.25  & 7.62  & 2.85  & 3.86  & 9.39  \\
 & BLEU\hspace{6ex}\hfill5 & 15.04  & 25.81  & 18.13  & 13.64  & 15.87  & \textbf{12.76} & 8.85  & 2.92  & 3.26  & 9.07  \\
 & 10  & \textbf{17.00} & \textbf{32.07} & \textbf{21.57} & 15.94  & \textbf{18.01} & 12.74  & \textbf{8.90}  & 3.16  & 4.00  & \textbf{9.92}  \\
 & doc2doc  & 12.18  & 26.38  & 12.43  & 14.53  & 13.55  & 9.98  & 6.55  & 2.73  & 2.47  & 8.15  \\ \cmidrule(lr){2-12}
 & 1 & 28.48  & 19.04  & 17.50  & 42.27  & 35.24  & 30.47  & 23.81  & 9.99  & \textbf{28.04} & 30.68  \\
 & 2 & 34.89  & 30.65  & 25.17  & 43.07  & 37.84  & 35.09  & 26.58  & 12.11  & 26.45  & 30.70  \\
 & chrF++\hspace{6ex}\hfill5 & 43.66  & 46.60  & 34.88  & 39.59  & 41.04  & 37.71  & 28.51  & 12.33  & 24.85  & 30.36  \\
 & 10  & \textbf{47.07} & \textbf{53.07} & \textbf{38.51} & \textbf{43.36} & \textbf{43.64} & \textbf{37.77} & \textbf{29.25} & 12.81  & 27.15  & \textbf{31.86} \\
 & doc2doc  & 40.09  & 47.84  & 27.44  & 41.77  & 37.72  & 33.17  & 26.40  & 11.86  & 24.68  & 28.94 \\
\bottomrule
\end{tabular}

% }
\caption{Results from LLMs fine-tuned with different chunk sizes.}
\label{tab:exp_5_3}
\end{table*}

\begin{table}[t]
\centering \small
\begin{tabular}{lcc}
\toprule
\multirow[b]{2}{*}
& \multicolumn{2}{c}{\textbf{Avg \#tokens per doc}}  \\
\cmidrule(lr){2-3}
 & \textbf{\dochplt} & \textbf{WMT24++} \\
\midrule
en-ar & 451 & 369 \\
en-ca & 550	& 402 \\
en-hi & 949	& 423 \\
en-et & 786	& 358 \\
en-fa & 582	& 397 \\
en-fi & 611 & 337 \\
en-is & 585	& 407 \\
en-ko & 602	& 338 \\
en-ml & 581	& 334 \\
en-ur & 822 & 448 \\
\bottomrule
\end{tabular}
\caption{Average whitespace-delimited tokens per English document in \dochplt{} and WMT24++ tests.}
\label{tab:avg_tokens_per_doc}
\end{table}

\subsection{How much context do document-level models need?} \label{4.3}
Existing research on DocMT with LLM adopts distinct strategies to process data. Some approaches operate on a sentence level but use previous translations as context \citep{wu2024adapting}, some methods process fixed-size chunks \citep{alabi2025afridoc,wicks-etal-2024-recovering,post-junczys-dowmunt-2024-evaluation}, translating each chunk separately, and some perform full document-to-document translation \citep{ramos2025multilingual}. We start our experiments by training models using varying context lengths to find the best configuration.

\paragraph{Setup} We build en-xx models for five target languages separately: Finnish, Icelandic, Korean, Malayalam, and Urdu. We fine-tune two open-source LLMs: Qwen2.5-7B-Instruct and Llama-3.1-8B-Instruct. We fine-tune each model under five different context configurations: sentence-level (chunk 1, no context), chunks of 2, 5, and 10 sentences, as well as full document-to-document (doc2doc) training. For chunk-based training, we compute the loss only on target segments while providing source context. The total number of tokens is kept constant for all languages, despite the different data formats.

\paragraph{Results}
Our experiments in Table~\ref{tab:exp_5_3} reveal a clear and consistent pattern on the \dochplt{} test set: measures of translation quality systematically improve as the input size increases from a single sentence to a 10-sentence chunk. As shown in the table, fine-tuning with 10-sentence chunks almost universally delivers the best performance across models, directions, and metrics. The gains are particularly dramatic for lower-resource pairs, such as en-is, where the BLEU score for Llama-3.1-8B-Instruct jumps from a sentence-level performance of 6.51 to 32.07. Nonetheless, full document-to-document training consistently underperforms the 10-sentence chunking strategy. This indicates that while substantial context is crucial, training LLMs on entire documents still poses challenges. This is consistent with \citet{peng2025investigatinglengthissuesdocumentlevel}'s findings that LLM-based translation degrades on longer documents.

\begin{table*}[t]
\centering\small
\begin{tabular}{rlrrrrrrrrrrr}
 \toprule
 &  &  & \multicolumn{5}{c}{\dochplt{}}  & \multicolumn{5}{c}{WMT24++} \\ \cmidrule(lr){4-8}\cmidrule(lr){9-13}
 & &  & \multicolumn{1}{l}{en-fi} & \multicolumn{1}{l}{en-is} & \multicolumn{1}{l}{en-ko} & \multicolumn{1}{l}{en-ml} & \multicolumn{1}{l}{en-ur} & \multicolumn{1}{l}{en-fi} & \multicolumn{1}{l}{en-is} & \multicolumn{1}{l}{en-ko} & \multicolumn{1}{l}{en-ml} & \multicolumn{1}{l}{en-ur} \\
 \midrule
\multirow{4}{*}{\makecell{Qwen2.5-\\7B-Instruct}} & \multirow{2}{*}{BLEU} & IT & 11.01 & 10.42 & 14.33 & 3.05  & 3.79  & 11.57 & 6.12  & 5.90 & 2.42  & 3.97           \\
 & & FT & \textbf{13.87} & \textbf{32.54} & \textbf{23.71} & \textbf{12.20}  & \textbf{11.11} & \textbf{12.20}  & \textbf{9.27}  & \textbf{6.37}  & \textbf{3.67}  & \textbf{5.75}  \\
\cmidrule(lr){2-13}
 & \multirow{2}{*}{chrF++} & IT & 43.6  & 31.85 & 32.08 & 22.9  & 23.36 & \textbf{40.76} & 27.28 & 19.57 & 23.19 & 24.31          \\
 & & FT & \textbf{44.01} & \textbf{53.74} & \textbf{40.87} & \textbf{37.23} & \textbf{34.72} & 38.71 & \textbf{30.46} & \textbf{20.9}  & \textbf{26.65} & \textbf{26.37} \\
 \midrule
\multirow{4}{*}{\makecell{Llama-3.1-\\8B-Instruct}} & \multirow{2}{*}{BLEU} & IT & 14.92 & 14.11 & 12.89 & 6.66  & 12.99 & 12.24 & 7.11  & \textbf{3.78}  & 3.03  & 8.67           \\
 & & FT & \textbf{17.00}  & \textbf{32.07} & \textbf{21.57} & \textbf{15.94} & \textbf{18.01} & \textbf{12.74} & \textbf{8.90} & 3.16  & \textbf{4.00}   & \textbf{9.92}  \\
\cmidrule(lr){2-13}
 & \multirow{2}{*}{chrF++} & IT & 46.42 & 38.45 & 30.67 & 32.59 & 39.40  & \textbf{38.96} & 27.88 & \textbf{14.71} & 25.32 & 30.71          \\
 & & FT & \textbf{47.07} & \textbf{53.07} & \textbf{38.51} & \textbf{43.36} & \textbf{43.64} & 37.77 & \textbf{29.25} & 12.81 & \textbf{27.15} & \textbf{31.86} \\
\bottomrule
\end{tabular}

\caption{Results from prompting instruction-tuned (IT) LLMs and those further fine-tuned (FT) on \dochplt{}.}
\label{tab:pt-ft-models-sorted}
\end{table*}

However, we cannot observe a clear trend for WMT24++ regarding the training context size. The results are inconsistent, and the benefits of larger context windows are less clear. In several cases, smaller context windows or even simple sentence-level fine-tuning outperform the larger-context models, such as Qwen2.5-7B-Instruct on en-ml and en-ur. We hypothesize that the performance difference is due to document length variation as a domain bias. WMT24++ documents have roughly half the average number of tokens compared to \dochplt{} (Table \ref{tab:avg_tokens_per_doc}), so most WMT24++ documents fit within a small chunk size. This creates a mismatch where training on larger chunk sizes is unnecessary or harmful, as longer contexts rarely occur in WMT24++. 

These results show that the optimal context strategy for document-level translation is not absolute but is dependent on the test data characteristics. Based on our findings, we establish a training chunk size of 10 for all subsequent experiments.

\subsection{Does fine-tuning on \dochplt{} help document-level translation?}
One key indicator of the usefulness of a data resource is whether practitioners can create better models using it. Although the origin of our data is web crawls, which may have been consumed by LLM pre-training, the parallelism signals are new in \dochplt{} and not accessible through pre-training. Thus, in this section, we
compare the results of fine-tuning LLMs to prompting baselines.

\paragraph{Setup} 
We compare our fine-tuned models with the best-performing data configuration of chunk size 10 to their corresponding off-the-shelf instruction-tuned models. Evaluation is done on held-out \dochplt{} test sets and WMT24++. 

\paragraph{Results}
Table \ref{tab:pt-ft-models-sorted} shows that fine-tuning on \dochplt{} produces notable improvements across nearly all settings, with gains inversely proportional to language resource levels. On \dochplt{} test, lower-resourced languages see bigger jumps, e.g., in BLEU: 10.42 to 32.54 for Icelandic, 3.05 to 12.20 for Malayalam, and 3.79 to 11.11 for Urdu, whereas gains are more modest for higher-resourced languages, e.g., 11.01 to 13.87 for Finnish. On WMT24++, baseline prompting performance is generally poor, often below 6 BLEU, and improvements from fine-tuning persist but are smaller in absolute terms. This may be attributed to WMT24++'s domain mismatch (e.g., social and speech) with \dochplt{}.

Our results suggest that off-the-shelf instruction-tuned models may already contain knowledge for these medium to low-resourced languages, yet fine-tuning on \dochplt{} consistently improves performance across these languages. This underscores the value of our \dochplt{}, which is the \textit{first} to cater to those languages in this task. Nonetheless, we note that a higher performance does not necessarily indicate higher data quality---it may also be a result of greater exposure to a given language or to document-level input and output. A causal analysis will be useful, but for most of the languages we study, there is no suitable alternative data to compare to at the moment.

\begin{table*}[t]
    \centering\small
    \begin{tabular}{r@{\hskip 2ex}r@{\hskip 2ex}rrrrrrrrrrr}
\toprule
\multicolumn{3}{r}{\multirow[b]{2}{*}{\textit{Seen Languages}}} & \multicolumn{5}{c}{\dochplt} & \multicolumn{5}{c}{WMT24++} \\
\cmidrule(lr){4-8}
\cmidrule(lr){9-13}
 & & & en-fi & en-is & en-ko & en-ml & en-ur & en-fi & en-is & en-ko & en-ml & en-ur \\
\cmidrule(lr){1-13}
\multirow{6}{*}{\makecell{Qwen2.5-\\7B-Instruct}} & \multirow{3}{*}{BLEU} & Mono\textsubscript{1K} & 13.87 & 32.54 & \textbf{23.71} & 12.20 & 11.11 & 12.20 & 9.27 & 6.37 & 3.67 & 5.75 \\
 & & Multi\textsubscript{1K} & 10.31 & 24.11 & 20.12 & 5.07 & 4.35 & 11.66 & 6.62 & \textbf{6.76} & 1.43 & 3.45 \\
 & & Multi\textsubscript{5K} & \textbf{14.05} & \textbf{35.13} & 23.60 & \textbf{14.70} & \textbf{13.07} & \textbf{13.42} & \textbf{10.02} & 6.62 & \textbf{4.18} & \textbf{7.70}  \\
\cmidrule(lr){2-13}
 & \multirow{3}{*}{chrF++} & Mono\textsubscript{1K} & \textbf{44.01} & 53.74 & \textbf{40.87} & 37.23 & 34.72 & 38.71 & 30.46 & 20.90 & 26.65 & 26.37 \\
 & & Multi\textsubscript{1K} & 38.01 & 44.70 & 37.56 & 26.07 & 22.32 & 37.56 & 26.40 & \textbf{21.50} & 18.44 & 21.12 \\
 & & Multi\textsubscript{5K} & \textbf{44.09} & \textbf{56.74} & 40.56 & \textbf{40.28} & \textbf{37.16} & 40.35 & \textbf{32.24} & 21.23 & \textbf{27.28} & \textbf{29.57} \\
\cmidrule(lr){1-13}
\multirow{6}{*}{\makecell{Llama-3.1-\\8B-Instruct}} & \multirow{3}{*}{BLEU} & Mono\textsubscript{1K} & \textbf{17.00} & 32.07 & \textbf{21.57} & 15.94 & \textbf{18.01} & 12.74 & \textbf{8.90} & 3.16 & \textbf{4.00} & 9.92 \\
 & & Multi\textsubscript{1K} & 13.57 & 25.75 & 17.58 & 10.15 & 13.24 & 11.23 & 7.03 & 3.24 & 2.83 & 7.62 \\
 & & Multi\textsubscript{5K} & 16.57 & \textbf{34.21} & 21.04 & \textbf{17.01} & 17.55 & \textbf{13.39} & 8.46 & 3.33 & 3.87 & \textbf{10.13} \\
\cmidrule(lr){2-13}
 & \multirow{3}{*}{chrF++} & Mono\textsubscript{1K} & \textbf{47.07} & 53.07 & \textbf{38.51} & 43.36 & \textbf{43.64} & 37.77 & \textbf{29.25} & 12.81 & \textbf{27.15} & \textbf{31.86} \\
 & & Multi\textsubscript{1K} & 43.04 & 47.64 & 34.74 & 36.55 & 37.88 & 36.07 & 26.04 & 12.63 & 24.33 & 27.31 \\
 & & Multi\textsubscript{5K} & 45.00 & \textbf{55.39} & 37.83 & \textbf{44.69} & 42.73 & 38.15 & 28.47 & 12.53 & 26.73 & 31.77 \\
\bottomrule
\end{tabular}
\caption{Results from monolingual and multilingual fine-tuning for \textit{seen languages}.}
\label{tab:mono_multi}
\vspace{3ex}
\end{table*}

\begin{table*}[t]
\centering\small
\begin{tabular}{r@{\hskip 2ex}r@{\hskip 2ex}rrrrrrrrrrr}
\toprule
\multicolumn{3}{r}{\multirow[b]{2}{*}{\textit{Unseen Languages}}} & \multicolumn{5}{c}{\dochplt} & \multicolumn{5}{c}{WMT24++} \\
\cmidrule(lr){4-8}
\cmidrule(lr){9-13}
 & & & en-et & en-ca & en-hi & en-fa & en-ar & en-et & en-ca & en-hi & en-fa & en-ar \\
\cmidrule(lr){1-13}
\multirow{6}{*}{\makecell{Qwen2.5-\\7B-Instruct}} & \multirow{3}{*}{BLEU} & IT & \textbf{7.39} & 26.47 & \textbf{11.40} & \textbf{8.34} & 13.63 & \textbf{7.82} & \textbf{19.41} & \textbf{9.87} & \textbf{10.14} & 8.65 \\
 & & Multi\textsubscript{1K} & 4.96 & \textbf{26.59} & 8.22 & 7.28 & \textbf{15.85} & 6.96 & 18.78 & 7.44 & 8.80 & \textbf{10.09} \\
 & & Multi\textsubscript{5K} & 4.74 & 25.41 & 7.01 & 4.21 & 14.28 & 6.48 & 18.06 & 7.43 & 4.96 & 9.74 \\
\cmidrule(lr){2-13}
 & \multirow{3}{*}{chrF++} & IT & \textbf{36.34} & \textbf{55.59} & \textbf{35.46} & \textbf{36.12} & 39.44 & \textbf{33.04} & \textbf{47.19} & \textbf{33.30} & \textbf{36.16} & 31.84 \\
 & & Multi\textsubscript{1K} & 26.85 & 54.92 & 27.74 & 31.83 & \textbf{42.15} & 28.02 & 45.17 & 26.47 & 31.67 & \textbf{34.04} \\
 & & Multi\textsubscript{5K} & 27.28 & 53.81 & 24.99 & 23.84 & 39.12 & 27.62 & 44.72 & 27.22 & 24.45 & 34.02 \\
\cmidrule(lr){1-13}
\multirow{6}{*}{\makecell{Llama-3.1-\\8B-Instruct}} & \multirow{3}{*}{BLEU} & IT & \textbf{11.50} & \textbf{32.19} & 22.13 & \textbf{13.26} & \textbf{12.62} & \textbf{9.20} & \textbf{20.82} & 12.58 & 9.50 & \textbf{6.94}  \\
 & & Multi\textsubscript{1K} & 8.95 & 31.45 & 23.08 & 12.65 & 11.33 & 8.29 & 19.60 & 12.66 & \textbf{9.60} & 6.37 \\
 & & Multi\textsubscript{5K} & 8.73 & 30.17 & \textbf{23.95} & 11.87 & 10.55 & 7.61 & 19.29 & \textbf{13.40} & 9.34 & 6.30 \\
\cmidrule(lr){2-13}
 & \multirow{3}{*}{chrF++} & IT & \textbf{41.88} & \textbf{57.73} & 47.56 & \textbf{40.80} & \textbf{37.51} & \textbf{32.87} & \textbf{44.46} & \textbf{35.44} & \textbf{32.68} & \textbf{28.26} \\
 & & Multi\textsubscript{1K} & 35.41 & 56.75 & 47.68 & 38.79 & 33.82 & 29.47 & 42.12 & 34.68 & 31.84 & 25.98 \\
 & & Multi\textsubscript{5K} & 34.08 & 55.63 & \textbf{47.96} & 37.76 & 32.44 & 28.05 & 41.83 & 35.35 & 31.04 & 25.19 \\
 \bottomrule
 \end{tabular}
    \caption{Results from prompting instruction-tuned (IT) LLMs and multilingual fine-tuning for \textit{unseen languages}.}
    \label{tab:tab:it_multi}
    \vspace{3ex}
\end{table*}

\subsection{Does multilingual training improve over monolingual models?}
While our monolingual fine-tuned models achieve significant gains over prompting baselines, deploying and maintaining separate models for each language presents scalability drawbacks. Furthermore, multilingual LLM fine-tuning may offer performance advantages over monolingual tuning \citep{chen-etal-2024-monolingual}. 
To test whether our multilingual \dochplt can be exploited for cross-lingual transfer in training, in this section, we build and assess multilingual models. Particularly, we test on both seen and unseen languages to determine whether benefits extend beyond training languages. 

\paragraph{Setup}
We compare three data configurations: a monolingual FT approach and two multilingual FT settings, resulting in three models:
\begin{itemize}
    \item Mono\textsubscript{1K}: a monolingual FT data approach that uses 1000 documents per language.
    \item Multi\textsubscript{1K}: a multilingual setting that uses 1000 documents combined, with 200 from each of the 5 languages, intended to match the total size for monolingual FT.
    \item Multi\textsubscript{5K}: another multilingual setting that uses 5000 documents in total, with 1000 from each language, intended to match the size for each language in monolingual FT.
\end{itemize}
All models are trained with consistent hyperparameters as in Appendix~\ref{appen:hyperparam}. We stick to our best-performing data configuration of chunk size 10. 

We evaluate those models on \dochplt{} and WMT24++ for all 5 training languages and 5 additional unseen languages: Arabic (ar), Catalan (ca), Estonian (et), Hindi (hi), and Persian (fa). These unseen languages are selected for their linguistic and/or script relation with the training languages. 

\paragraph{Results}

Table \ref{tab:mono_multi} compares multilingual to monolingual FT, showing that multilingual advantages are model-dependent. For Qwen2.5-7B-Instruct, Multi\textsubscript{5K} outperforms Mono\textsubscript{1K} and Multi\textsubscript{1K} consistently, whereas Llama-3.1-8B-Instruct displays a mixed pattern. Taking a closer look at the languages, Icelandic and Malayalam always improve with multilingual training, regardless of the LLM. In Table~\ref{tab:tab:it_multi}, for unseen languages, we see that the off-the-shelf IT models are usually better than multilingual fine-tuning.

In general, multilingual fine-tuning produces inconsistent results: it improves DocMT performance over monolingual fine-tuning for some LLMs, but we find almost no zero-shot cross-lingual transfer. Our observations from a small-scale multilingual experiment warrant further investigation by scaling the model choices and sizes, as well as the number of languages which is supported by \dochplt{}.

\FloatBarrier
\section{Conclusion}
We introduced a pipeline to derive a document-level corpus with rich metadata and presented the outcome, \dochplt{}, the largest publicly available document-level translation dataset with 124 million aligned document pairs across 50 languages paired with English. The utility of our massively multilingual dataset has been demonstrated through experiments: fine-tuning LLMs on our data improved over prompting baselines, and multilingual training surpassed monolingual models, though zero-shot transfer to unseen languages remained challenging. Our experiments also revealed challenges in DocMT: full document-to-document training and generalization to other document domains. Future work may use \dochplt{} data for further investigations such as large-scale training, data filtering, data synthesis, and DocMT metric study. 

\section*{Acknowledgements}
\lettrine[image=true, lines=2, findent=1ex, nindent=0ex, loversize=.15]{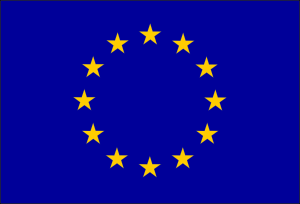}%
This project has received funding from the European Union's Horizon Europe research and innovation programme under grant agreement No 101070350 and from UK Research and Innovation (UKRI) under the UK government’s Horizon Europe funding guarantee [grant number 10052546]. 

Dayyán O’Brien is also supported by a G-Research NextGen Scholarship, part of the UKRI AI Centre for Doctoral Training in Responsible and Trustworthy in-the-world Natural Language Processing (grant ref: EP/Y030656/1).

We acknowledge the EuroHPC Joint Undertaking for awarding this project access to the EuroHPC supercomputer LUMI, hosted by CSC (Finland) and the LUMI consortium through a EuroHPC Regular Access call.

\bibliography{anthology.min.bib,custom.bib}

\appendix

\begin{table}[htbp]
\section{Dataset Statistics} \label{appen:data_stats}
\vspace{2ex}
\centering\small
\begin{tabular}{lrrr}
\toprule
 & \textbf{\#sentences} & \textbf{\#docs} & \textbf{\#deduped docs} \\
\midrule
af & 16,416,841 & 297,636 & 286,861 \\
ar & 65,482,300 & 2,271,167 & 2,196,334 \\
az & 12,202,189 & 332,742 & 321,500 \\
be & 10,672,952 & 212,121 & 203,758 \\
bg & 80,018,549 & 1,746,301 & 1,669,696 \\
bn & 10,473,372 & 414,099 & 405,339 \\
bs & 20,635,243 & 514,615 & 488,093 \\
ca & 47,905,003 & 1,198,217 & 1,131,468 \\
cy & 8,908,119 & 265,261 & 253,040 \\
en & 2,665,945,834 & 47,484,349 & 45,995,228 \\
eo & 6,115,355 & 119,196 & 103,858 \\
et & 33,684,509 & 774,561 & 747,075 \\
eu & 6,783,654 & 189,347 & 175,318 \\
fa & 24,837,952 & 810,029 & 785,963 \\
fi & 111,615,913 & 2,445,791 & 2,341,993 \\
ga & 6,398,081 & 172,167 & 166,516 \\
gl & 10,657,570 & 233,545 & 215,009 \\
gu & 3,202,679 & 108,507 & 106,423 \\
he & 38,077,820 & 1,190,198 & 1,149,349 \\
hi & 37,592,475 & 1,336,090 & 1,315,174 \\
hr & 52,267,826 & 1,063,347 & 1,009,227 \\
is & 12,571,982 & 274,078 & 265,088 \\
ja & 164,136,152 & 4,032,689 & 3,934,457 \\
kk & 5,948,866 & 140,082 & 135,689 \\
kn & 4,463,262 & 123,053 & 120,996 \\
ko & 84,527,642 & 2,058,811 & 2,003,338 \\
lt & 48,692,264 & 1,031,628 & 995,288 \\
lv & 37,426,957 & 796,659 & 766,138 \\
mk & 12,465,228 & 307,055 & 292,992 \\
ml & 2,925,457 & 115,189 & 111,721 \\
mr & 3,066,703 & 128,808 & 126,552 \\
ms & 51,150,528 & 978,185 & 942,418 \\
mt & 6,328,544 & 141,088 & 137,104 \\
nb & 89,189,502 & 1,884,362 & 1,809,266 \\
ne & 1,549,852 & 74,579 & 73,691 \\
nn & 4,228,079 & 93,285 & 78,426 \\
si & 1,497,375 & 50,605 & 48,730 \\
sk & 70,057,465 & 1,461,804 & 1,406,726 \\
sl & 37,501,647 & 797,858 & 765,171 \\
sq & 11,475,561 & 328,651 & 317,315 \\
sr & 21,620,629 & 407,440 & 386,829 \\
sw & 8,409,824 & 185,287 & 178,185 \\
ta & 6,790,864 & 215,564 & 208,573 \\
te & 5,131,680 & 141,279 & 138,727 \\
th & 16,134,265 & 676,699 & 655,628 \\
tr & 100,380,235 & 3,884,137 & 3,767,266 \\
uk & 89,841,883 & 1,955,041 & 1,891,287 \\
ur & 5,479,098 & 234,708 & 228,952 \\
uz & 3,502,356 & 69,440 & 68,191 \\
vi & 87,511,126 & 1,986,258 & 1,940,095 \\
xh & 995,556 & 21,561 & 20,797 \\
\midrule
\textbf{total} & \textbf{4,264,894,818} & \textbf{87,775,169} & \textbf{84,882,858} \\
\bottomrule
\end{tabular}
\caption{A summary of \dochplt{} documents and sentences per language.}
\label{tab:language_totals}
\end{table}

\begin{table*}[htbp]
\centering\small
\resizebox{0.99\textwidth}{!}{
\begin{tabular}{lrrccrrccc}
\toprule
& \multicolumn{1}{r}{\multirow[b]{2}{*}{\textbf{\#doc pairs}}} & \multicolumn{1}{r}{\multirow[b]{2}{*}{\textbf{\#alignments}}} & \multirow[b]{2}{*}{\makecell{\textbf{avg \#aligns.}\\\textbf{/\#doc}}} & \multirow[b]{2}{*}{\makecell{\textbf{avg  \#sents\_en}\\\textbf{/\#sents\_xx}}} & \multicolumn{2}{c}{\textbf{\#sents/\#docs}} & \multirow[b]{2}{*}{\makecell{\textbf{avg}\\\textbf{BLEUalign}}} & \multirow[b]{2}{*}{\makecell{\textbf{avg}\\\textbf{Bicleaner}}} & \multirow[b]{2}{*}{\makecell{\textbf{avg align.}\\\textbf{density}}}\\
\cmidrule(lr){6-7}
 & & & & & \multicolumn{1}{c}{\textbf{en}} & \multicolumn{1}{c}{\textbf{xx}} &  & & \\
\midrule
af-en & 1,121,166 & 29,496,715 & 26.3 & 1.38 & 85.1 & 108.5 & 0.551 & 0.418 & 0.446 \\
ar-en & 4,405,876 & 54,747,241 & 12.4 & 0.84 & 35.7 & 56.6 & 0.468 & 0.700 & 0.280 \\
az-en & 732,657 & 10,289,514 & 14.0 & 1.26 & 53.1 & 64.7 & 0.443 & 0.482 & 0.334 \\
be-en & 709,129 & 14,728,785 & 20.8 & 1.37 & 87.7 & 104.9 & 0.543 & 0.556 & 0.324 \\
bg-en & 6,016,906 & 93,051,525 & 15.5 & 1.34 & 65.9 & 79.9 & 0.541 & 0.582 & 0.285 \\
bn-en & 1,039,423 & 7,851,362 & \phantom{0}7.6 & 0.89 & 34.4 & 69.8 & 0.446 & 0.577 & 0.182 \\
bs-en & 1,443,819 & 17,704,604 & 12.3 & 1.20 & 65.4 & 95.6 & 0.512 & 0.516 & 0.268 \\
ca-en & 3,582,267 & 63,520,169 & 17.7 & 1.20 & 60.4 & 87.6 & 0.562 & 0.620 & 0.335 \\
cy-en & 721,671 & 12,632,309 & 17.5 & 1.15 & 45.4 & 60.9 & 0.577 & 0.618 & 0.426 \\
eo-en & 482,452 & 8,677,590 & 18.0 & 3.61 & 147.6 & 82.1 & 0.511 & 0.474 & 0.246 \\
et-en & 2,484,493 & 40,019,712 & 16.1 & 1.96 & 74.5 & 56.7 & 0.502 & 0.501 & 0.311 \\
eu-en & 616,924 & 8,245,785 & 13.4 & 2.85 & 88.4 & 52.0 & 0.493 & 0.402 & 0.294 \\
fa-en & 1,880,900 & 11,884,837 & \phantom{0}6.3 & 2.69 & 76.2 & 40.1 & 0.423 & 0.544 & 0.153 \\
fi-en & 8,532,601 & 135,452,163 & 15.9 & 1.80 & 76.0 & 61.9 & 0.546 & 0.555 & 0.307 \\
ga-en & 557,716 & 10,060,287 & 18.0 & 1.85 & 61.2 & 49.6 & 0.613 & 0.488 & 0.419 \\
gl-en & 988,176 & 15,903,011 & 16.1 & 3.06 & 120.3 & 69.4 & 0.533 & 0.532 & 0.256 \\
gu-en & 306,386 & 3,358,243 & 11.0 & 2.65 & 91.6 & 52.7 & 0.476 & 0.500 & 0.187 \\
he-en & 4,190,235 & 49,247,941 & 11.8 & 2.91 & 85.7 & 40.8 & 0.537 & 0.577 & 0.220 \\
hi-en & 3,502,520 & 32,907,313 & \phantom{0}9.4 & 2.55 & 60.9 & 36.5 & 0.479 & 0.609 & 0.196 \\
hr-en & 3,574,689 & 54,626,216 & 15.3 & 1.77 & 82.0 & 72.3 & 0.537 & 0.528 & 0.302 \\
is-en & 1,097,797 & 19,959,668 & 18.2 & 2.02 & 77.4 & 52.6 & 0.498 & 0.474 & 0.333 \\
ja-en & 11,828,819 & 144,978,567 & 12.3 & 1.75 & 63.7 & 49.9 & 0.462 & 0.382 & 0.181 \\
kk-en & 243,579 & 4,197,879 & 17.2 & 1.47 & 72.9 & 60.7 & 0.446 & 0.559 & 0.381 \\
kn-en & 355,117 & 5,270,814 & 14.8 & 2.65 & 124.1 & 71.2 & 0.446 & 0.515 & 0.200 \\
ko-en & 6,479,547 & 106,313,693 & 16.4 & 1.98 & 78.9 & 54.7 & 0.526 & 0.572 & 0.237 \\
lt-en & 3,948,829 & 62,315,769 & 15.8 & 1.78 & 74.5 & 64.5 & 0.538 & 0.546 & 0.283 \\
lv-en & 3,104,028 & 53,107,619 & 17.1 & 1.96 & 77.8 & 63.9 & 0.555 & 0.573 & 0.308 \\
mk-en & 961,749 & 17,710,874 & 18.4 & 2.03 & 94.8 & 65.8 & 0.518 & 0.576 & 0.319 \\
ml-en & 298,334 & 2,211,378 & \phantom{0}7.4 & 3.91 & 89.6 & 36.9 & 0.427 & 0.459 & 0.151 \\
mr-en & 372,093 & 2,567,437 & \phantom{0}6.9 & 3.44 & 70.4 & 33.4 & 0.432 & 0.446 & 0.150 \\
ms-en & 3,887,463 & 69,632,512 & 17.9 & 2.01 & 82.2 & 65.6 & 0.551 & 0.390 & 0.289 \\
mt-en & 477,497 & 9,464,200 & 19.8 & 1.72 & 69.7 & 59.7 & 0.605 & 0.293 & 0.407 \\
nb-en & 6,596,166 & 105,226,440 & 16.0 & 1.61 & 66.7 & 58.8 & 0.542 & 0.556 & 0.308 \\
ne-en & 201,928 & 1,415,859 & \phantom{0}7.0 & 3.03 & 57.8 & 26.1 & 0.448 & 0.394 & 0.169 \\
nn-en & 413,279 & 4,396,370 & 10.6 & 3.56 & 113.0 & 55.9 & 0.445 & 0.421 & 0.164 \\
si-en & 123,803 & 1,338,609 & 10.8 & 2.36 & 83.5 & 51.9 & 0.474 & 0.442 & 0.219 \\
sk-en & 5,262,604 & 81,849,513 & 15.6 & 1.56 & 73.6 & 66.0 & 0.536 & 0.597 & 0.293 \\
sl-en & 2,334,208 & 41,082,011 & 17.6 & 1.73 & 80.2 & 68.7 & 0.503 & 0.536 & 0.329 \\
sq-en & 910,599 & 15,055,014 & 16.5 & 1.93 & 78.0 & 58.6 & 0.529 & 0.515 & 0.382 \\
sr-en & 1,307,126 & 25,315,953 & 19.4 & 1.88 & 106.8 & 78.4 & 0.541 & 0.492 & 0.335 \\
sw-en & 581,466 & 12,214,107 & 21.0 & 1.95 & 98.2 & 84.0 & 0.557 & 0.340 & 0.348 \\
ta-en & 583,034 & 5,804,724 & 10.0 & 2.68 & 84.5 & 41.8 & 0.458 & 0.434 & 0.190 \\
te-en & 389,858 & 5,202,332 & 13.3 & 2.83 & 123.4 & 68.4 & 0.466 & 0.495 & 0.178 \\
th-en & 2,438,548 & 18,656,911 & \phantom{0}7.7 & 2.76 & 57.5 & 30.5 & 0.501 & 0.531 & 0.197 \\
tr-en & 11,815,778 & 120,528,089 & 10.2 & 2.80 & 62.4 & 34.5 & 0.520 & 0.503 & 0.215 \\
uk-en & 5,364,321 & 88,197,354 & 16.4 & 1.64 & 80.8 & 68.5 & 0.516 & 0.608 & 0.312 \\
ur-en & 618,996 & 5,471,488 & \phantom{0}8.8 & 2.94 & 70.2 & 39.1 & 0.463 & 0.508 & 0.198 \\
uz-en & 156,796 & 3,300,674 & 21.1 & 1.61 & 85.2 & 76.7 & 0.461 & 0.492 & 0.369 \\
vi-en & 5,089,734 & 66,322,073 & 13.0 & 1.72 & 76.2 & 61.2 & 0.413 & 0.626 & 0.235 \\
xh-en & 44,001 & 1,276,014 & 29.0 & 1.67 & 96.3 & 101.3 & 0.477 & 0.443 & 0.407 \\
\midrule
Average  & 2,483,542 & 35,495,785 & 14.8 & 2.11 & 79.4 & 61.8 & 0.503 & 0.510 & 0.277 \\
Total & 124,177,103 & 1,774,789,267 &  &  &  &  &  &  &  \\
\bottomrule
\end{tabular}
}
\caption{A summary of \dochplt{} alignment statistics by language pair.}
\label{tab:alignment_stats}
\end{table*}

\clearpage
\section{Hyperparamters} \label{appen:hyperparam}
Below, we list the hyperparameters used during training.
\begin{table}[h!]
\centering\small
\begin{tabular}{ll}
\toprule
\textbf{Parameter} & \textbf{Value} \\
\midrule
Learning Rate & 5e-04 \\
LR Scheduler Type & Linear \\
Warmup Ratio & 0.1 \\
Weight Decay & 0.0 \\
\addlinespace
Per Device Train Batch Size & 2 \\
Gradient Accumulation Steps & 4 \\
Number of Train Epochs & 1 \\
\addlinespace
LoRA Rank & 16 \\
LoRA Alpha & 32 \\
\addlinespace
Seed & 1729 \\
\bottomrule
\end{tabular}
\caption{Training Hyperparameters}
\label{tab:hyperparameters}
\end{table}

\section{Prompts} \label{appen:prompts}
\subsection{Overview}
We use the same prompt for SFT and during inference for both off-the-shelf instruction-tuned and fine-tuned models. LLM's chat template is always applied.

We illustrate chunk-based translation and full document-to-document translation using the task of English to Catalan translation.

\subsection{Chunk-based translation}

\textbf{Template (chunk size 2):}
\begin{tcolorbox}[colback=gray!10, colframe=gray!50, boxrule=0.5pt]
\small
Translate the following source segment from [SOURCE LANGUAGE] into [TARGET LANGUAGE].\\[0.5em]
[SOURCE LANGUAGE]: [SOURCE TEXT]\\[0.5em]
[TARGET LANGUAGE]: [TARGET TEXT]
\end{tcolorbox}

\noindent\textbf{Example:}
\begin{tcolorbox}[colback=blue!5, colframe=blue!30, boxrule=0.5pt]
\small
Translate the following source segment from English into Catalan.\\[0.5em]
English: Online workshops are organized every month. 
The results will be shared with the community.\\[0.5em]
Catalan: Cada mes s'organitzen tallers en línia. Els 
resultats es compartiran amb la comunitat.
\end{tcolorbox}

\subsection{Document-to-document translation}

\textbf{Template}
\begin{tcolorbox}[colback=gray!10, colframe=gray!50, boxrule=0.5pt]
\small
Translate the following source document from [SOURCE LANGUAGE] into [TARGET LANGUAGE].\\[0.5em]
[SOURCE LANGUAGE]: [SOURCE DOCUMENT]\\[0.5em]
[TARGET LANGUAGE]: [TARGET DOCUMENT]
\end{tcolorbox}

\noindent\textbf{Example:}
\begin{tcolorbox}[colback=blue!5, colframe=blue!30, boxrule=0.5pt]
\small
Translate the following source document from English into Catalan.\\[0.5em]
English: Our proposals with you in mind. We suggest.... 
Castelló d'Empúries is situated in the heart of the Aiguamolls 
Natural Park. Stay at a house in the historic center of Castelló 
d' Empúries Check opening times and escape from the hustle and 
bustle of the city with a visit you will love. A weekend to 
explore Empordà by bike. Here you will also find events, fairs 
and festivals that are held close to Hostal Casa Clara.\\[0.5em]
Catalan: Les nostres propostes pensades per a vosaltres. 
Us suggerim... Castelló d'Empúries està situat al bell mig del 
Parc Natural dels Aiguamolls. Les teves vacances en una casa al 
centre històric de Castelló d'Empúries Consulta els horaris i 
fes una visita que t'encantarà i et farà desconnectar del brogit 
de ciutat. Un cap de setmana en bicicleta per conèixer l'Empordà. 
Aquí també hi trobaràs esdeveniments, fires i festes populars 
que es fan prop de l'Hostal Casa Clara
\end{tcolorbox}

\end{document}